%% file: cvpr.tex
\documentclass[final]{cvpr}

\usepackage{times}
\usepackage{epsfig}
\usepackage{graphicx}
\usepackage{amsmath}
\usepackage{amssymb}

\usepackage{booktabs}
\usepackage{xcolor,colortbl}
\definecolor{Gray}{gray}{0.9}
\usepackage{multirow}
\usepackage{caption}
\usepackage{subcaption}
\usepackage{nopageno}

\newcommand\blfootnote[1]{%
  \begingroup
  \renewcommand\thefootnote{}\footnote{#1}%
  \addtocounter{footnote}{-1}%
  \endgroup
}

\usepackage[pagebackref=true,breaklinks=true,colorlinks,bookmarks=false]{hyperref}

\def\cvprPaperID{3034} 
\def\confYear{CVPR 2021}

\begin{document}

\title{Generalizing Face Forgery Detection with High-frequency Features}


\author{
Yuchen Luo$^{*\dagger1,2}$
\qquad Yong Zhang$^{*3}$
\qquad Junchi Yan$^{\ddagger1,2}$ 
\qquad Wei Liu$^{\ddagger4}$\\
$^{1}$Department of Computer Science and Engineering, Shanghai Jiao Tong University \\
$^{2}$MoE  Key  Lab  of  Artificial  Intelligence,  AI Institute, Shanghai Jiao Tong University\\
$^{3}$Tencent AI Lab
\qquad $^{4}$Tencent Data Platform\\
{\tt\small 
\{592mcavoy,yanjunchi\}@sjtu.edu.cn
\qquad zhangyong201303@gmail.com
\qquad wl2223@columbia.edu}
}


\maketitle

\input{tex/0.abstract}

\blfootnote{$^*$Equal contribution}
\blfootnote{$^\dagger$Work done during an internship at Tencent AI Lab}
\blfootnote{$^\ddagger$Corresponding Author}
\input{tex/1.introduction}
\input{tex/2.relatedwork}
\input{tex/3.analysis}
\input{tex/4.approach}
\input{tex/5.experiment}
\input{tex/6.conclusion}

\input{tex/acknowledge}

{\small
\bibliographystyle{ieee_fullname}
\bibliography{fake}
}

\end{document}

%% file: tex/0.abstract.tex
\begin{abstract}


Current face forgery detection methods achieve high accuracy under the within-database scenario where training and testing forgeries are synthesized by the same algorithm. 
However, few of them gain satisfying performance under the cross-database scenario where training and testing forgeries are synthesized by different algorithms.
In this paper, we find that current CNN-based detectors tend to overfit to method-specific color textures and thus fail to generalize.
Observing that image noises remove color textures and expose discrepancies between authentic and tampered regions, we propose to utilize the high-frequency noises for face forgery detection.
We carefully devise three functional modules to take full advantage of the high-frequency features.
The first is the multi-scale high-frequency feature extraction module that extracts high-frequency noises at multiple scales and composes a novel modality.
The second is the residual-guided spatial attention module that guides the low-level RGB feature extractor to concentrate more on forgery traces from a new perspective.
The last is the cross-modality attention module that leverages the correlation between the two complementary modalities to promote feature learning for each other.
Comprehensive evaluations on several benchmark databases corroborate the superior generalization performance of our proposed method.
\end{abstract}
\vspace{-20pt}

%% file: tex/1.introduction.tex
\section{Introduction}
 

As face manipulation techniques~\cite{faceaap,deeperforensics10,Zhu2020AOTAO} spring up along with the breakthrough of deep generators~\cite{kingma2013auto, goodfellow2014generative}, face forgery detection becomes an arousing research topic.
Most existing methods focus on within-database detection~\cite{agarwal2019protecting,li2018ictu,amerini2019deepfake}, where forged images in the training set and testing set are manipulated by the same algorithm.
However, the biggest challenge hampering face forgery detection is the generalization problem.
Due to the diversified data distributions generated by different manipulation techniques, methods with high within-database detection accuracy always experience a severe performance drop in the cross-database scenario, thus limiting broader applications.

Recently, several works are devoted to addressing the generalizing problem.
\cite{li2018exposing,li2019face} assume that some artifacts are shared in forged faces and customize databases specialized on those artifacts.
However, these assumptions do not always hold.
Besides, transfer learning~\cite{cozzolino2018forensictransfer}, domain adaptation~\cite{yang2020one}, and multi-task learning~\cite{du2019towards,nguyen2019multi,stehouwer2019detection} are utilized to improve model's performance in unseen domains.
Nevertheless, the acquisition of target samples and annotations is expensive.
Meanwhile, some attempt to obtain information from frequency domains, such as Fourier transformation~\cite{chen2020manipulated}, DCT~\cite{f3net}, and steganalysis features~\cite{wu2020sstnet,zhou2017two}. 
But they rarely consider the relation and interaction between the additional information and regular color textures.

\input{figure/our-work}

In this paper, we aim at learning a more generalizable face forgery detector (See Fig.~\ref{fig:our-work}).
To facilitate understanding why current CNN-based works fail on unseen forgeries, we analyze CNN-based classifiers' behaviors and find that the model is biased to method-specific color textures.
Observing that high-frequency noises can suppress image textures and expose statistical discrepancies between tampered and authentic regions, we propose to utilize noises to tackle the overfitting problem.


We propose a generalizable model for face forgery detection.
To take full advantage of image noises, we carefully devise three novel modules.
The first is the \textit{multi-scale high-frequency feature extraction module}.
We adopt the widely used high-pass filters from SRM~\cite{fridrich2012rich} to extract high-frequency noises from images.
Unlike~\cite{chen2020manipulated,wu2020sstnet,zhou2017two} that only consider extracting noises from an input image, we further apply these filters to low-level features at multiple scales to compose more abundant and informative features.  
Employing both the high-frequency noises and the low-frequency textures, we build a two-stream network to process the two modalities, respectively.
Secondly, we apply the \textit{residual guided spatial attention} in the entry part to guide the RGB modality from the residual perspective to attach more importance to forgery traces.
Thirdly, we design a \textit{dual cross-modality attention module} to formulate the interaction between the two modalities instead of keeping them independent~\cite{chen2020manipulated,wu2020sstnet}.
In this way, the two modalities provide complementary information based on their correlation and mutually promote representation learning.


Our contributions are summarized as follows: 
\begin{itemize}
    \item We perform an analysis on CNN-based detectors and find that they are biased to method-specific textures, leading to the generalization problem. 
    Given that the image's high-frequency noises can remove color textures and reveal forgery traces, we propose to utilize image noises to boost the generalization ability. 
    \item We devise a generalizable model by exploiting high-frequency features and modeling the correlation and interaction between the high-frequency modality and the regular one. 
    We design three functional modules for learning an adequate representation, \textit{i.e.}, the multi-scale high-frequency feature extraction module, the residual guided spatial attention module, and the dual cross-modality attention module.  
    \item We conduct comprehensive evaluations on several benchmarks and demonstrate the superior generalization ability of the proposed model.
\end{itemize}

%% file: figure/our-work.tex
\begin{figure}[t]
  \centering
  \includegraphics[width=\linewidth]
  {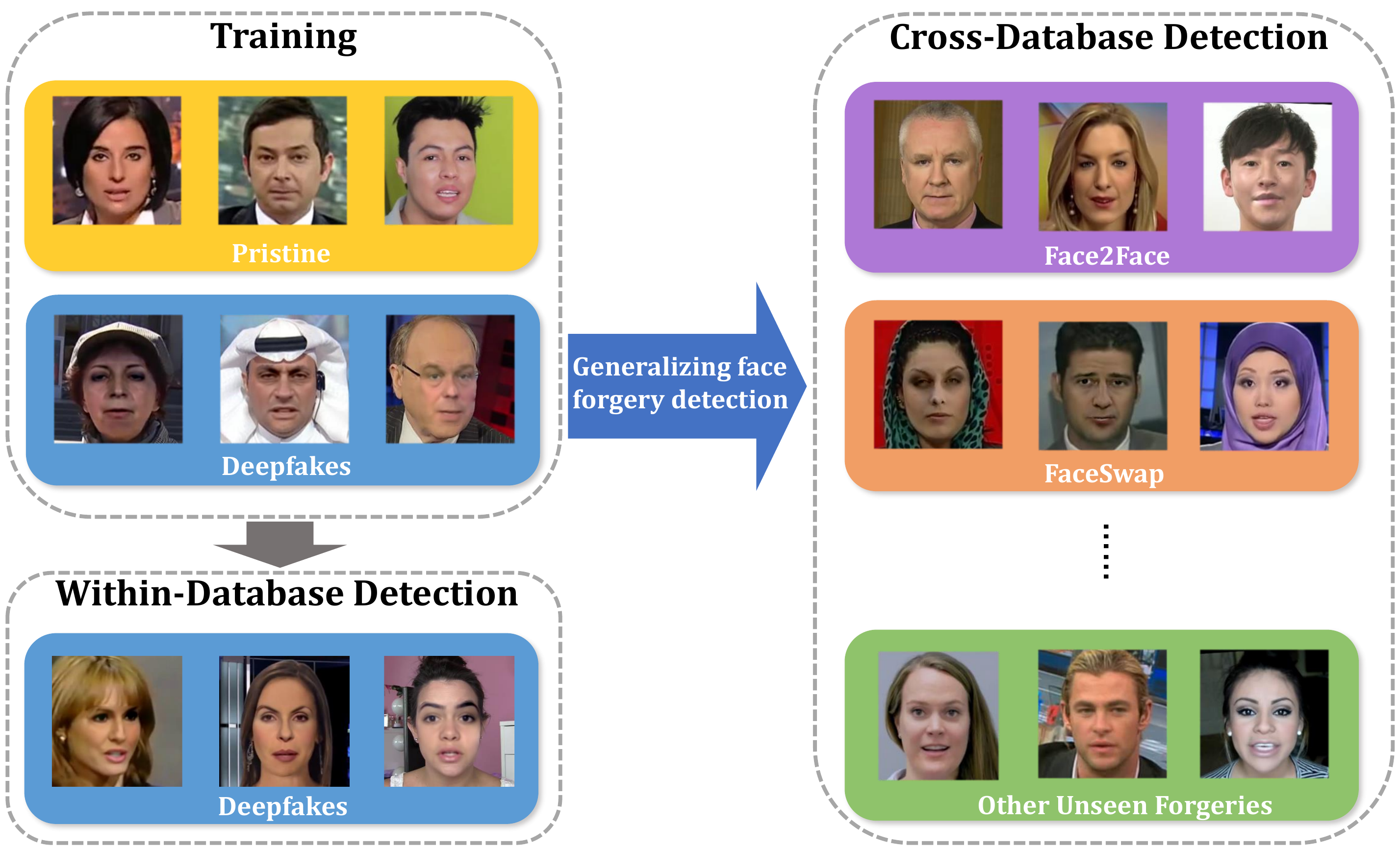} 
  \caption
    {
    Training and testing forgeries of within-database detection are synthesized by the same algorithm while those of cross-database detection are synthesized by different algorithms. 
    We focus on the latter which is more challenging. 
    }
 \label{fig:our-work}
 \vspace{-15pt} 
\end{figure}

%% file: tex/2.relatedwork.tex
\section{Related Work}
\input{figure/grad-cam}



\paragraph{Conventional image forgery detection.}
Though image tampering detection has been investigated for a long time, those conventional detectors~\cite{farid2009image,rocha2011vision} cannot handle the detection of face forgery well.
Firstly, those methods detect image editing operations like copy-move or copy-paste~\cite{amerini2011sift}, which produce different artifacts from those in GAN-based face forgeries.
Besides, face forgeries have a much smaller size and inferior quality than natural images.
Furthermore, recent advanced forgery techniques leave almost no visible artifact in the tampered face, which easily deceive conventional detectors and require specialized treatments.


\vspace{-8pt}

\paragraph{Specific artifacts or novel architectures.}
Current attempts mainly focus on improving within-database performance.
\cite{li2018ictu,matern2019exploiting,yang2019exposing,amerini2019deepfake,agarwal2019protecting, gram-net} targeted at specific artifacts such as abnormal eye-blinking frequency~\cite{li2018ictu} or head-pose inconsistency~\cite{yang2019exposing}.
However, those artifacts may not exist in improved forgeries.
For network design, Afchar \textit{et al.}~\cite{afchar2018mesonet} provided two compact network architectures to capture the mesoscopic features. 
Nguyen \textit{et al.}~\cite{nguyen2019capsule} introduced the capsule network.
These methods emphasize the power of representation and computational efficiency but did not explicitly consider the generalization ability.  
\vspace{-8pt}

\paragraph{Auxiliary tasks or synthetic data.}
Some works notice the generalization problem and attempt to utilize additional tasks~\cite{du2019towards,nguyen2019multi,stehouwer2019detection,songsri2019complement} 
or generate samples capturing the typical defects~\cite{li2018exposing,li2019face,yang2020one}.
Stehouwer~\textit{et al.}~\cite{stehouwer2019detection} proposed to jointly predict the binary label and the attention map of the manipulated region. 
Yang \textit{et al.}~\cite{yang2020one} generated samples with similar distributions to those in the target domain.
These methods require additional annotations or extra data samples.
Li \textit{et al.}~\cite{li2018exposing} observed that forged images contain some common warping and blurring effects.
Li \textit{et al.}~\cite{li2019face} focused on the fusion operation in forgery creation.
Both of them customized a database to learn such artifacts, but there remain non-negligible gaps between the hand-crafted forgeries and those from sophisticated algorithms.

\vspace{-8pt}

\paragraph{High-frequency features.}
Several methods tried to exploit information from other domains.
Durall \textit{et al.}~\cite{durall2019unmasking} found that the spectrum of real and fake images distributes differently in the high-frequency part.
Chen~\textit{et al.}~\cite{chen2020manipulated} proposed a multi-stream design and leveraged the DFT features.
Wu \textit{et al.}~\cite{wu2020sstnet} proposed a two-stream network to extract spatial features and steganalysis features, respectively. 
Qian \textit{et al.}~\cite{f3net} applied DCT on images and collected the frequency-aware clues to mine subtle forgery artifacts and compression errors.
These methods exploited frequency information but did not explicitly consider the relationship between different domains.


%% file: figure/grad-cam.tex
\begin{figure*}[t]
  \centering
 \includegraphics[width=0.80\linewidth]{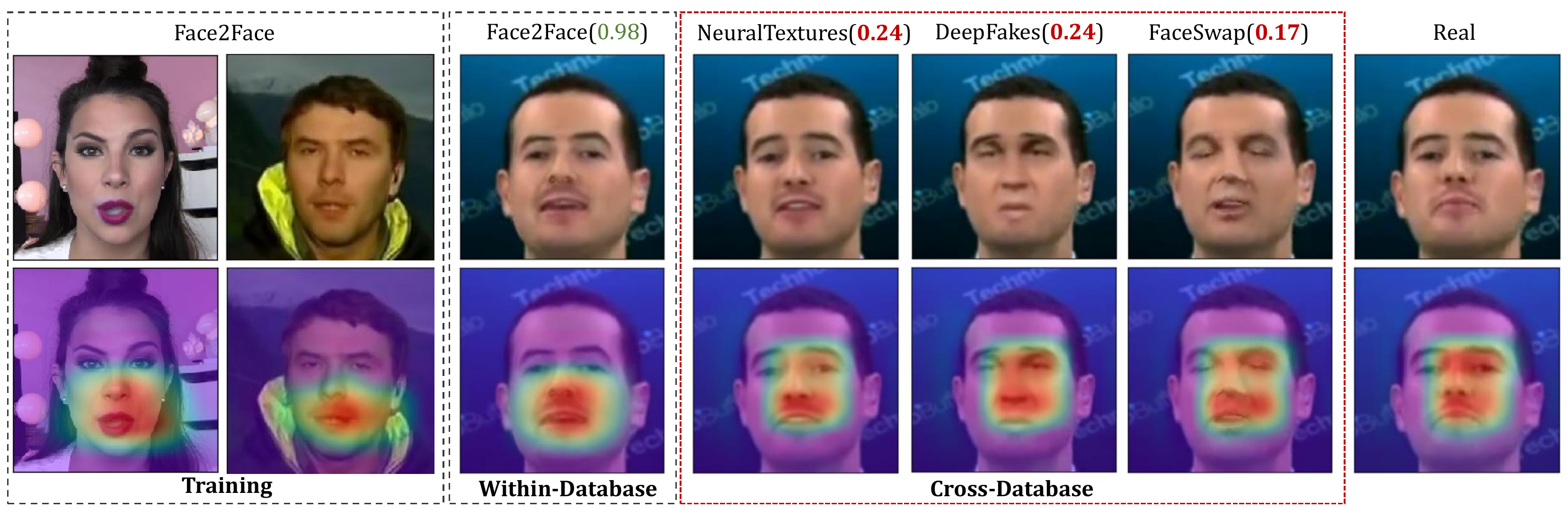}
  \caption
  {
  Grad-CAM maps from the Xception model trained on F2F forgeries.
  Numbers in the bracket denote the probability of being classified as fake. 
The mouth region is especially highlighted in F2F images, indicating that the model learns F2F's specific texture.
  But when evaluated on unseen forgeries (\textit{i.e.,} NT, DF, and FS), the model's responses resemble those in the real face more, showing that it fails to recognize identical artifacts and mistakes these forgeries as real.
  }
 \label{fig:gradcam}
  \vspace{-0.4cm} 
\end{figure*}


%% file: tex/3.analysis.tex
\section{Analysis of Generalizable Forgery Detector}



\subsection{Why current methods fail to generalize?}


\input{table/xception-generalization}
\paragraph{CNN and texture bias.} 
Although many previous methods perform flawlessly on the testing set, they always suffer from a significant performance drop on images manipulated by unseen algorithms (See Tab.~\ref{tab:xception-bad-gen}).
Why do they fail to generalize?
The reason is that those deep CNN models learn to capture the method-specific texture patterns for forgery detection~\cite{gram-net}. 
Geirhos \textit{et al.}~\cite{geirhos2018imagenettrained} studied the texture response of CNNs and showed that CNN models are strongly biased to textures.
Different forgery algorithms always have unique network architectures and processing streams, so images manipulated by different algorithms will have different \textit{fake textures}.
Therefore, it is hard for a CNN model that has already been biased to one kind of fake textures to generalize to another.
\vspace{-8pt}

\paragraph{Grad-CAM study and results.}
To verify that the CNN classifier is biased to specific fake textures, we exploit the gradient-based visualization tool~\cite{grad-cam} to generate class activation maps, namely Grad-CAM maps, which reveal the regions that CNNs rely on for classification.
We train Xception~\cite{8099678} on images forged by the Face2Face method and then evaluate it on forgeries of four different algorithms.



As presented in Fig.~\ref{fig:gradcam},
in the within-database evaluation, the model correctly finds out forgeries with high confidence.
It especially concentrates on regions around the mouth (warmer color region), just as it does on the training images.
Nevertheless, when encountering unseen forgeries, the model mistakes the other three kinds of fakes as real though they contain many visible artifacts.
Besides, the high response regions deviate significantly from those in the training set and resemble those in the real face.
The reason is that the model has overfitted to F2F's unique fake textures.
Thus, when it cannot recognize the identical fake texture patterns, it always gives wrong predictions.

The above observation motivates us to investigate more general clues other than the texture patterns.

\subsection{What is common in forged face images?}

\input{figure/man-pipeline}

\paragraph{Manipulation pipeline.}
To find the commonality among different manipulation methods, let us review the typical face manipulation pipeline firstly. 
As presented in Fig.~\ref{fig:man-pipeline}, the manipulation procedure can be roughly divided into two processing stages, \textit{i.e.}, \textit{fake face creation} and \textit{face blending}.
In the first stage, fake faces are generated by sophisticated deep networks~\cite{thies2019deferred, deepfake} or rendered based on model templates~\cite{thies2016face2face, faceswap}. 
Different manipulation methods always adopt different algorithms and thus produce varied texture patterns.
In the second stage, the generated face is further refined and warped back to the original image.
This stage usually consists of some post-processing operations such as color correction and image blending.
\vspace{-8pt}

\paragraph{Discrepancy brought by forged part.}
Though the face region in the output image is manipulated, the background remains the same as in the source image (See Fig.~\ref{fig:man-pipeline}).
Assuming that different images have unique characteristics, the blending stage violates the original data distributions, and we can utilize the characteristics discrepancy to generalize forgery detection. 
We share a similar intuition to \cite{li2019face}, which focuses on the blending operation and detects the blending boundary.
But instead of identifying the \textit{boundary pattern} which may change with various post-processing operations, we expect a more robust way to discover the inconsistency between authentic and tampered regions.


\input{figure/srm-example}
\subsection{Use SRM to extract high-frequency noise features and boost the generalization ability}

Based on the above analysis, we assume that a generalizable forgery detector should 
(i) pay attention not only to texture-related features but also to texture-irrelevant ones,
and 
(ii) be capable of discovering the discrepancy between the tampered face and pristine background.
Observing that the image's high-frequency noises remove the color content while portrait the intrinsic characteristics, we attempt to utilize image noises for face forgery detection.

\vspace{-8pt}
\paragraph{Image Noise Analysis.}
Noises are some high-frequency signals capturing random variations of brightness or color information.
The distributions of image noises are influenced by the image sensor or circuitry of a digital camera.
Hence images processed by different equipment or coming from different sources have different noise patterns.
Noises can be viewed as an intrinsic specificity of an image and can be found in various forms in all digital imagery domains~\cite{noise-1}.
Given that manipulation operations ruin the consistency of characteristics in the original image, there often leave distinctive traces in the noise space~\cite{noise-2,noise-3}.

\vspace{-8pt}
\paragraph{SRM for noise feature extraction.}
Inspired by recent progress on SRM~\cite{fridrich2012rich} noise features in general image manipulation detection~\cite{zhou2018learning,wu2019mantra}, we adopt SRM filters for noise extraction.
Fig.~\ref{fig:srm-example} shows some examples of SRM noises.

To validate noise features' effectiveness, we compare the detection performance on RGB color textures and SRM noises features, respectively.
We train two Xception models, one with the regular RGB images and the other with the SRM noises.
The two models are trained on images forged by the F2F method and evaluated on all four methods.
As presented in the first two rows in Tab.~\ref{tab:ablation}, the model with SRM noise generalizes better than that with the regular color textures, especially on FS and NT methods.


%% file: table/xception-generalization.tex
\begin{table}[t]
  \centering
  \caption{Xception~\cite{8099678} detector experiences a significant performance drop when evaluated on unseen forgeries.}
  \scalebox{0.80}{
  \begin{tabular}{c|cccc}
    \toprule 
    \multirow{2}*{Training Set}
    &\multicolumn{4}{c}{Testing Set (AUC)} \\
    \cmidrule(lr){2-5}
    ~& DF& F2F& FS& NT \\ 
    \midrule
    DF  &\cellcolor{Gray}0.993	&0.736	&0.485	&0.736 \\
    F2F &0.803	&\cellcolor{Gray}0.994	&0.762	&0.696 \\
    FS  &0.664	&0.888	&\cellcolor{Gray}0.994	&0.713 \\
    NT  &0.800	&0.813	&0.731	&\cellcolor{Gray}0.991 \\
    
    \bottomrule
  \end{tabular}
  }
  
  \label{tab:xception-bad-gen}
  \vspace{-0.4cm} 
\end{table}

%% file: figure/man-pipeline.tex
\begin{figure}[t]
  \centering
 \includegraphics[width=0.80\linewidth]{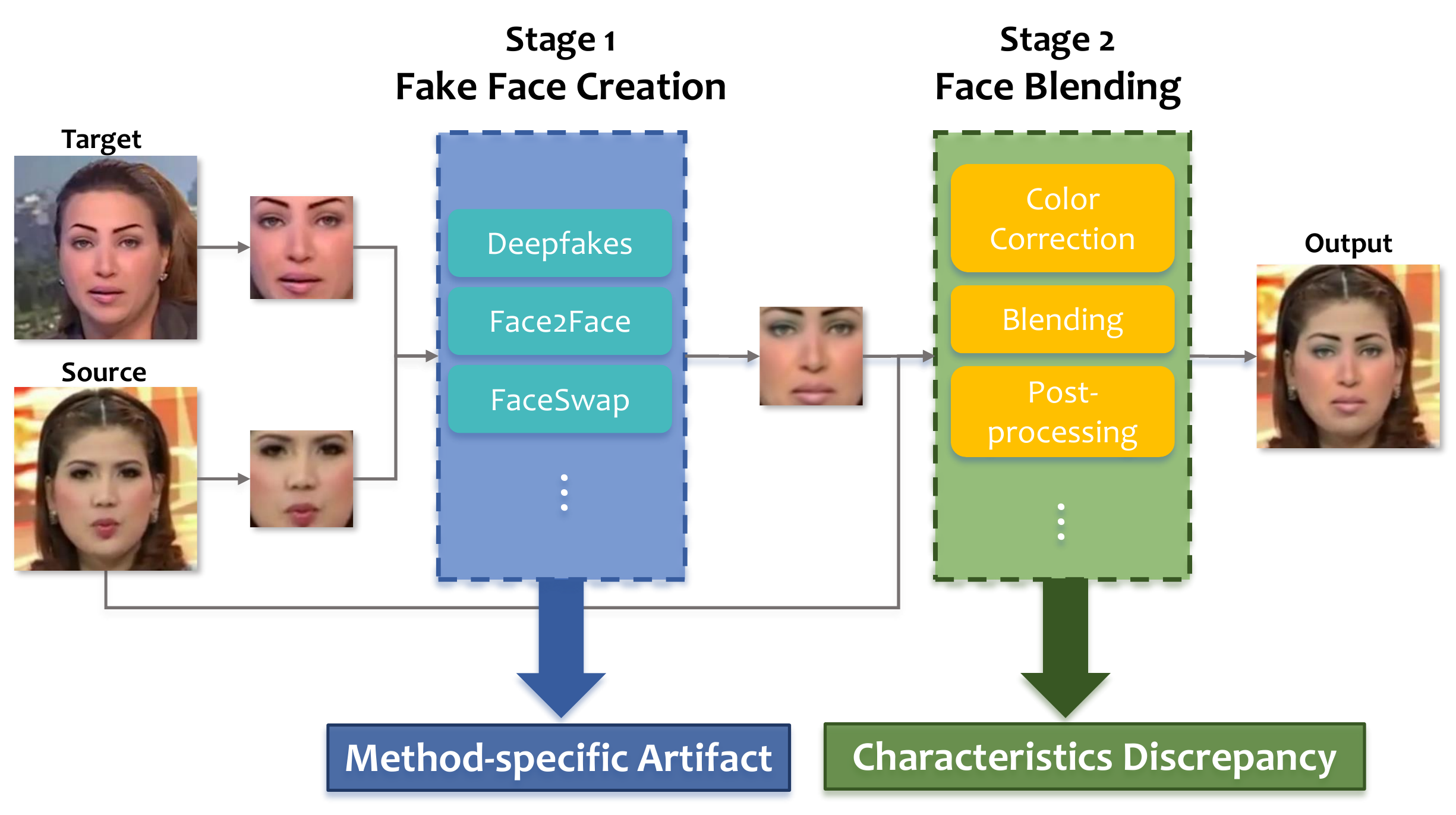}
  \caption{The overview of the typical manipulation pipeline. 
  The two stages bring the method-specific texture artifacts and characteristics discrepancy, respectively. }
 \label{fig:man-pipeline}
  \vspace{-0.45cm} 
\end{figure}

%% file: figure/srm-example.tex

\begin{figure}[t]
  \centering
 \includegraphics[width=0.85\linewidth]{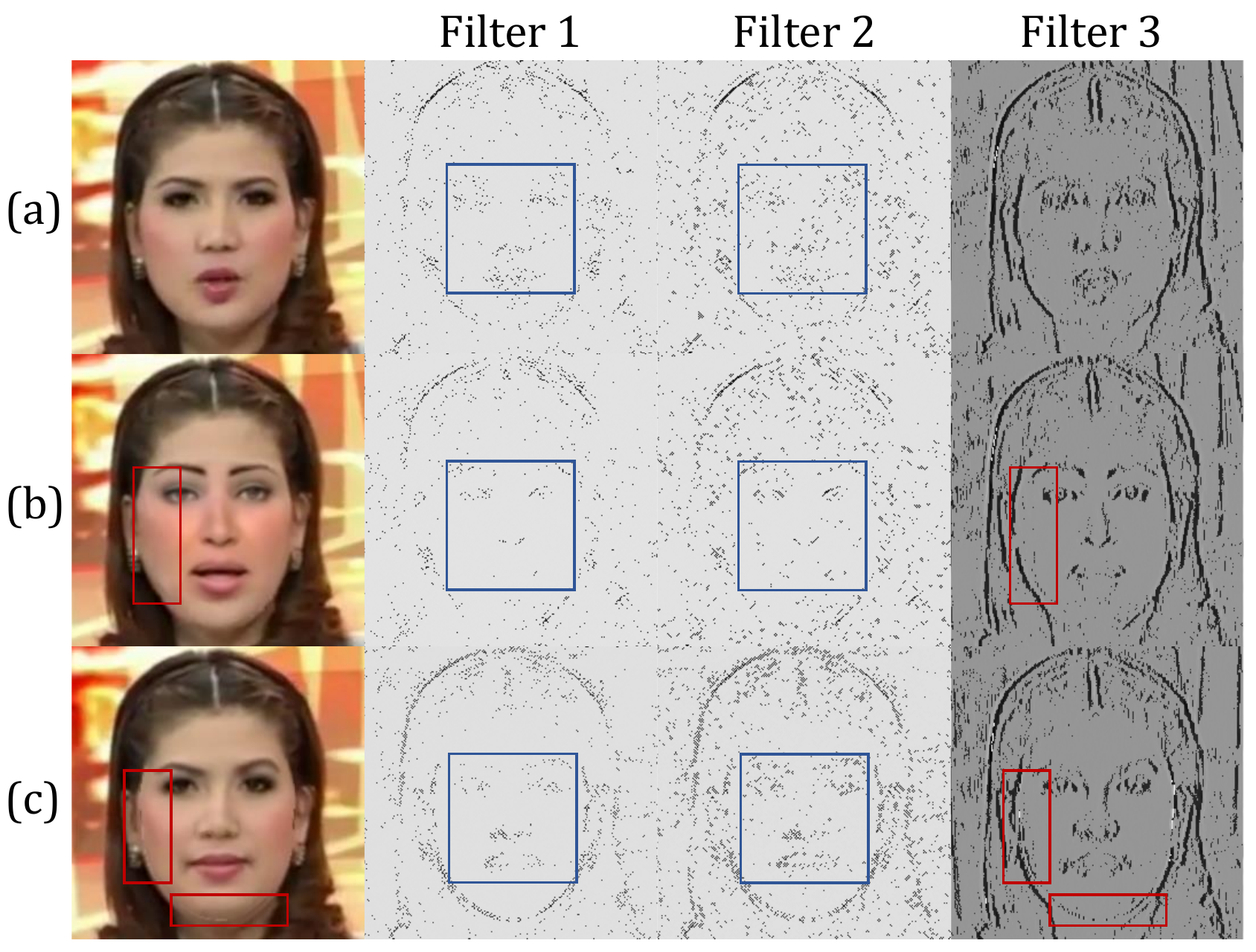}
  \caption{
    Noises extracted by SRM from (a) authentic face and face manipulated by (b) Deepfakes and (c) Face2Face.
    Red boxes mark blending traces that are hard to recognize in the RGB space but distinctive in the noise space.
    Blue boxes mark noise statistics in the central face region.
    Noises in the face region distribute continuously in authentic images but are visibly smoother or sharper in tampered images.
    }
 \label{fig:srm-example}
  \vspace{-0.45cm} 
\end{figure}


%% file: tex/4.approach.tex
\section{The Proposed Method}
The promising performance of SRM noises motivates us to explore the noise space further and boost generalization.
In this section, we devise three modules to take full advantage of the high-frequency features, \textit{i.e.,} a multi-scale high-frequency feature extraction module, a dual cross-modality attention module, and a residual guided spatial attention module.  
The proposed model is illustrated in Fig.~\ref{fig:pipeline}.
\input{figure/pipeline}

\subsection{Multi-scale High-frequency Feature Extraction}
Previous methods~\cite{zhou2017two,wu2019mantra,chen2020manipulated} extract noise residuals solely from the input image.
Apart from a straight-forward conversion of the input image, we apply high-pass filters to multiple low-level feature maps to enrich the high-frequency features.
As shown in Fig.~\ref{fig:pipeline}, given an input RGB image $\mathbf{X}$, we convert it to a residual image $\mathbf{X}_h$ in the high-frequency domain exploiting the SRM filters~\cite{fridrich2012rich}. 
The entry flow takes both $\mathbf{X}$ and $\mathbf{X}_h$ as input and generates two types of raw features, \textit{i.e.,} the multi-scale high-frequency feature maps $\mathbf{F}_h$ and the low-frequency spatial feature maps $\mathbf{F}$,
which can be formulated as 
\begin{align} \label{entry}
    \mathbf{F}, \mathbf{F}_h &= f_{\text{entry}}(\mathbf{X}, \mathbf{X}_h).
\end{align}

The multi-scale high-frequency features are obtained as follows.
Firstly, we apply regular convolutions on $\mathbf{X}$ and $\mathbf{X}_h$ to produce feature maps $\mathbf{F}^1$ and $\mathbf{F}_h^1$.  
To extract more high-frequency information, we then apply SRM filters on $\mathbf{F}^1$, following a $1\times 1$ convolution to align the channel dimensions and get the output $\Tilde{\mathbf{F}}_h^1$.
As $\mathbf{F}_h^1$ and $\Tilde{\mathbf{F}}_h^1$ are obtained from different sources and different operations, they contain different information.
We down-sample $\mathbf{F}_1$ as well as the sum of $\mathbf{F}_h^1$ and $\Tilde{\mathbf{F}}_h^1$, and then get $\mathbf{F}^2$ and $\mathbf{F}_h^2$ in the two streams, respectively.
Repeating the above operations,
we finally acquire the multi-scale high-frequency feature maps $\mathbf{F}_h$. 
Compared with $\mathbf{X}_h$, $\mathbf{F}_h$ embodies more abundant high-frequency signals from both the image and low-level feature maps at multiple scales.

\input{figure/srm_sa}

\subsection{Residual Guided Spatial Attention}
Inspired by CBAM~\cite{woo2018cbam}, we adopt spatial attention to highlight the manipulation traces and guide the feature learning in the RGB modality.
While CBAM derives the attention weights from the RGB image, we exploit the residuals generated by the SRM filters to predict the attention maps, which guide from a high-frequency perspective (See Fig.~\ref{fig:srm_sa}).
Considering that the spatial correspondence between the noise residuals and low-level RGB features is still well maintained, we place several spatial attention modules in the entry part.
The residual guided spatial attention block is defined as 
\begin{equation}
    \mathbf{M} = f_{\text{att}} (\mathbf{X}_h), 
\end{equation}
where $\mathbf{M}$ is the output attention map and $\mathbf{X}_h$ is the residual image.
The raw feature maps $\mathbf{F}$ in Eq.~(\ref{entry}) are computed by feeding the element-wise production $\mathbf{M} \odot \mathbf{F}^2$ into consequent convolutions and down-sampling operations. 

See the attention visualizations in Fig.~\ref{fig:att-example}, high responses occur around abnormal facial boundaries in manipulated faces but distribute uniformly in real ones, which implies that the residual guided spatial attention can help feature extractor focus on forgery traces.


\input{figure/dual_cma}

\subsection{Dual Cross-modality Attention}
The attention mechanism has been broadly applied in natural language processing~\cite{vaswani2017attention} and computer vision~\cite{hu2018squeeze, pedersoli2017areas, Huang_2019_ICCV}. 
As for the cross-modality attention, Ye \textit{et al.}~\cite{ye2019cross} applied self-attention to concatenated features, while Hou \textit{et al.}~\cite{hou2019cross} derives correlation-based attention for paired class and query features.
Inspired by these works, we devise a dual cross-modality attention module (DCMA) to capture long-range dependency and model the interaction between
the low-frequency textures and the high-frequency noises.

Denoting input features derived from the RGB stream and the high-frequency stream as $\mathbf{T}$ and $\mathbf{T}_h$, respectively, the DCMA module leverages the computation block described in Fig.~\ref{fig:dual_cma} to convert them to $\mathbf{T}'$ and $\mathbf{T}_h'$:
\begin{align}
    \mathbf{T}', \mathbf{T}_h' = f_{\text{DCMA}} (\mathbf{T}, \mathbf{T}_h). 
\end{align}

Taking the conversion of $\mathbf{T}$ as an example, we first convert the input $\mathbf{T} \in \mathbb{R}^{C\times H \times W}$ into two components through different convolution blocks.
One value component $\mathbf{V} \in \mathbb{R}^{C \times H \times W}$ represents the domain-specific information, 
the other key component $\mathbf{K}\in \mathbb{R}^{C/r \times H \times W}$ measures the correlation between two different domains. 
$r$ is a scalar that reduces $\mathbf{K}$'s channel dimension for computation efficiency.
Input features $\mathbf{T}_h \in \mathbb{R}^{C\times H \times W}$ in the high-frequency domain are converted into two components, \textit{i.e.,} $\mathbf{V}_h$ and $\mathbf{K}_h$, similarly. 
Note that the key component is obtained through a two-layer convolution block.
We set the first layer independently while sharing the second layer to project feature maps from the two modalities onto the same space.

Then we measure the correlation $\mathbf{C}\in \mathcal{R}^{HW\times HW}$ between the two modalities through $\mathbf{C} = \text{flt}(\mathbf{K})^T \otimes \text{flt}(\mathbf{K}_h)$, where $\text{flt}(\cdot)$ denotes the flatten operation and $\otimes$ denotes matrix multiplication. 
For the RGB attention, we multiply the correlation $\mathbf{C}$ with weight matrix $\mathbf{W}$ and generate the attention map $\mathbf{A}$ through the softmax operation, which is formulated as $\mathbf{A} =  \text{softmax}(\mathbf{C} \otimes \mathbf{W})$.
Similarly, we obtain the high-frequency attention map $\mathbf{A}_h = \text{softmax}(\mathbf{C}^T \otimes \mathbf{W}_h)$.
$\mathbf{A}$ and $\mathbf{A}_h$ re-weight features in one modality according to its correlation with the other.

Applying $\mathbf{A}$ to the corresponding value component $\mathbf{V}_h$, we obtain the refined features $\mathbf{R}$, where $\mathbf{R} = \text{flt}(\mathbf{V}_h) \otimes \mathbf{A}$.
We then recover $\mathbf{R}$ to the input dimension, and output the features $\mathbf{T}'$ of the RGB stream by $\mathbf{T}' = \mathbf{T} + \mathbf{R}$.
The calculation for features $\mathbf{T}_h'$ in the high-frequency stream is the same.
$\mathbf{T}'$ and $\mathbf{T}_h'$ embody complementary information and promote the feature learning for each other.
As shown in the pipeline, the DCMA module can be applied $N$ times to model the correlations between the two modalities at different scales.
We set $N=2$ in our experiments. 

\vspace{-8pt}
\paragraph{Feature Fusion and Loss Function.}
High-level features of the two modalities are fused at the end of the exit flow (See Fig.~\ref{fig:pipeline}).
We apply channel-wise attention~\cite{hu2018squeeze} on the concatenated features and then make the prediction.
Inspired by progress in face recognition~\cite{cosface, facereg},
we adopt the AM-Softmax Loss~\cite{wang2018additive} as the objective function since it leads to smaller intra-class variations and larger inter-class differences than the regular cross-entropy loss.

%% file: figure/pipeline.tex
\begin{figure*}[t]
  \centering
 \includegraphics[width=0.85\linewidth]{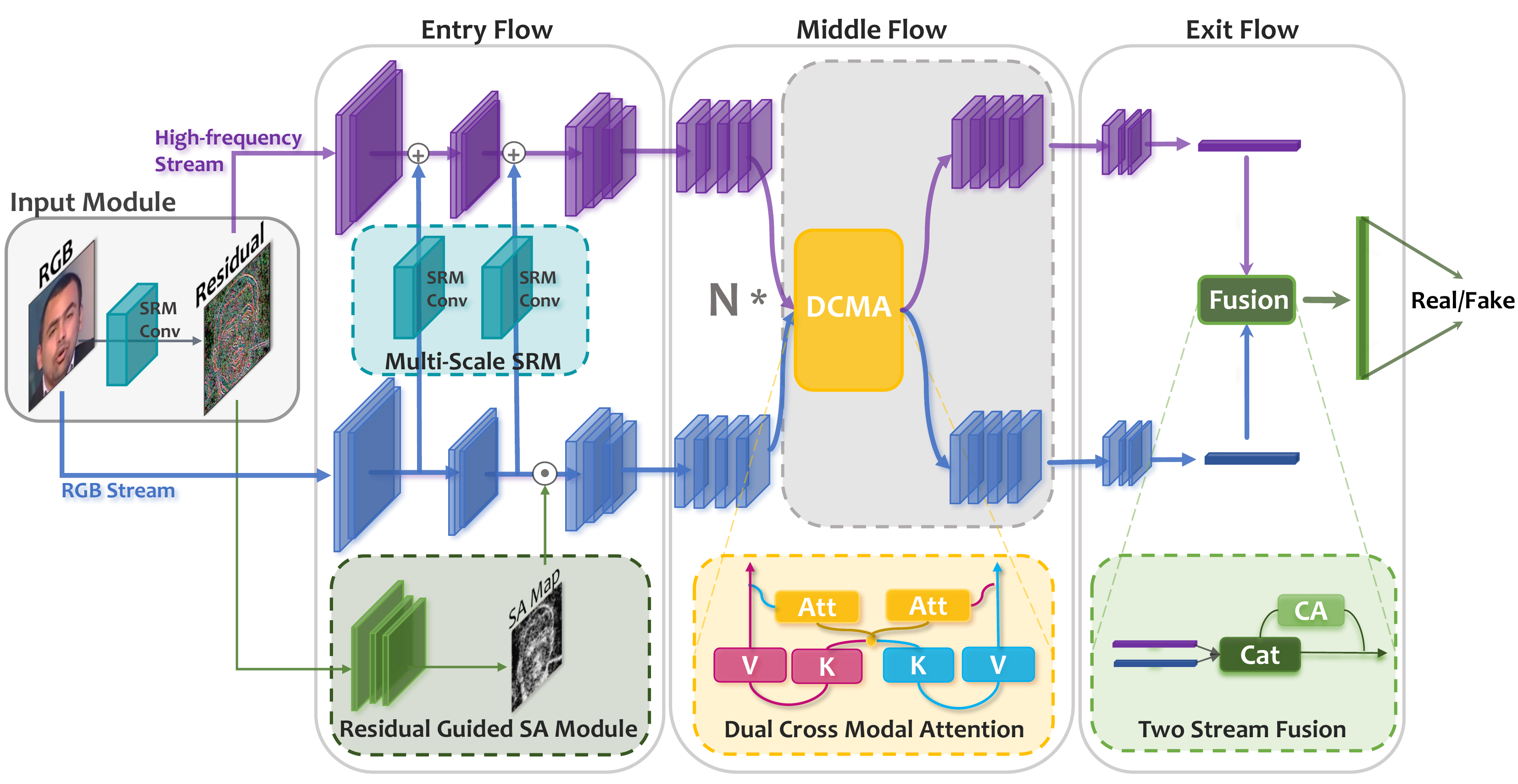}
  \caption{
  The pipeline of the proposed model. 
  We design a two-stream architecture to process the RGB image and the high-frequency noises. 
  In the entry flow, we extract multi-scale high-frequency features and residual guided spatial features. 
  In the middle flow, we model the interaction between feature maps of the two modalities via several DCMA modules.
  Features from the two streams are fused in an attention-based manner for the final classification.}
 \label{fig:pipeline}
  \vspace{-0.4cm} 
\end{figure*}

%% file: figure/srm_sa.tex
\begin{figure}[t]
    \centering
    \scalebox{0.8}{
    \includegraphics[width=\linewidth]{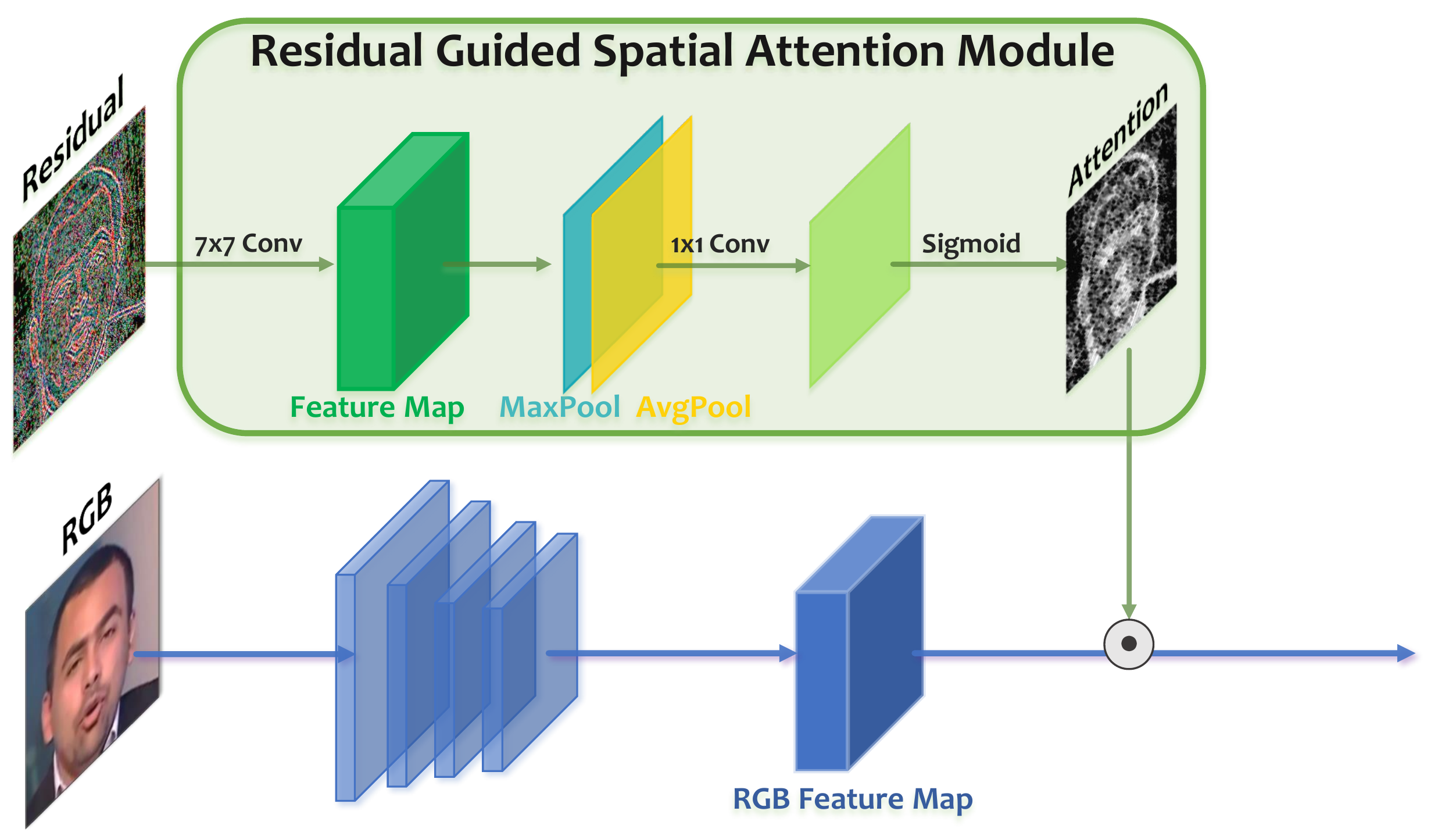}}
    \caption{The residual guided spatial attention.}
    \label{fig:srm_sa}
     \vspace{-0.4cm} 
\end{figure}

%% file: figure/dual_cma.tex
\begin{figure}[t]
    \centering
    \scalebox{0.8}{
    \includegraphics[width=\linewidth]{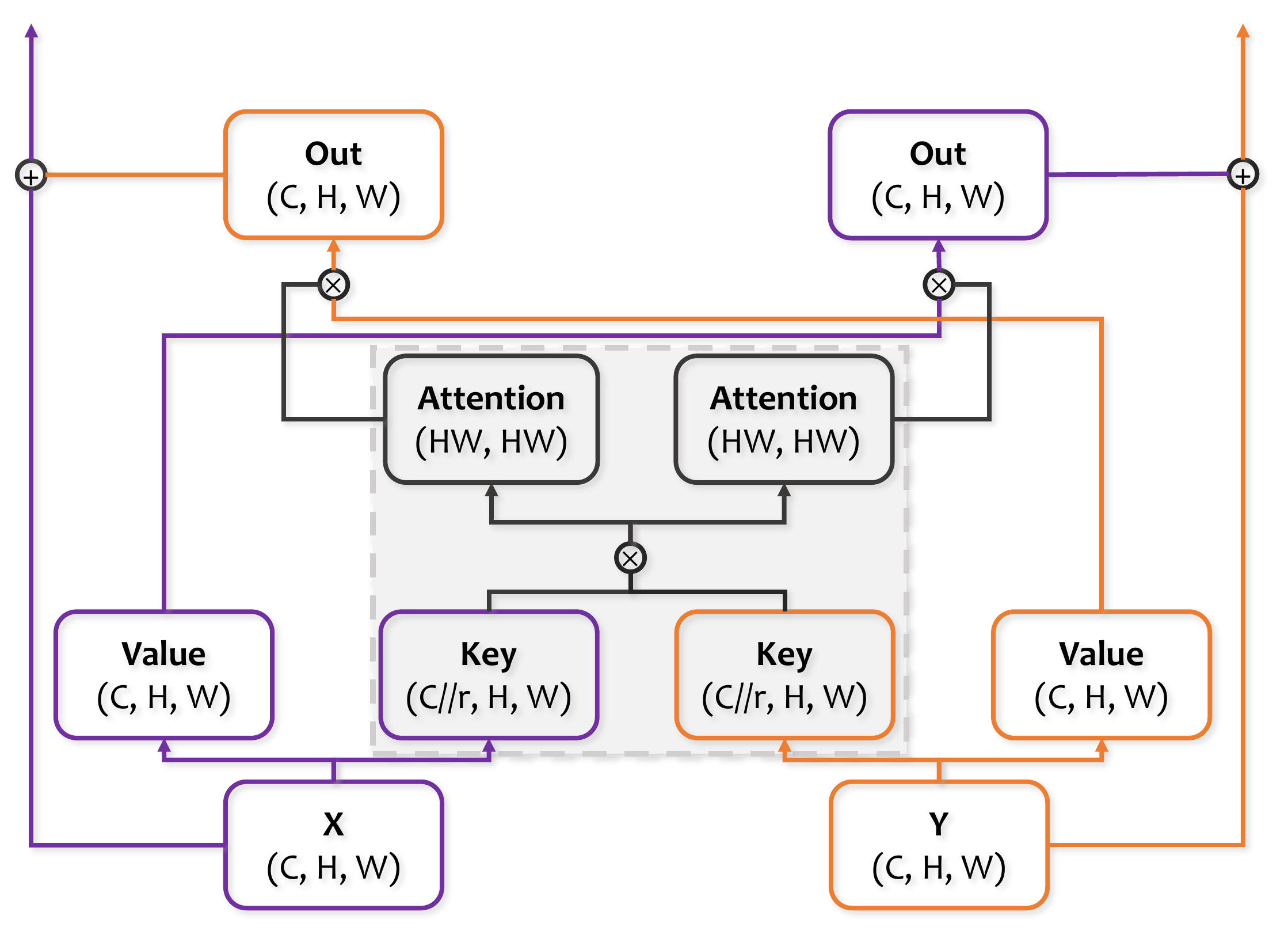}}
    \caption{The dual cross-modality attention.}
    \label{fig:dual_cma}
     \vspace{-0.4cm} 
\end{figure}

%% file: tex/5.experiment.tex
\input{figure/att_example}

\section{Experiments}

\subsection{Settings}
\paragraph{Datasets.}
To evaluate the generalization ability, we perform experiments on five large scale benchmark databases, \textit{i.e.,} FaceForensics++ (FF++)~\cite{rossler2019faceforensics++}, DeepfakeDetection (DFD)~\cite{dfd}, Deepfake Detection Challenge (DFDC)~\cite{dfdc}, CelebDF~\cite{li2019celeb}, and DeeperForensics-1.0 (DF1.0)~\cite{deeperforensics10}. 
Detailed specifications are presented in Tab.~\ref{tab:datasets}.
For evaluation on FF++, we follow the official splits by using 740 videos for training, 140 videos for validation, and 140 videos for testing.
There are three versions of FF++ in terms of compression level, \textit{i.e.}, raw, lightly compressed (HQ), and heavily compressed (LQ).
The heavier the compression level, the harder it to distinguish the forgery traces.
Since realistic forgeries always have a limited quality, we use the HQ and LQ versions in experiments.
\textbf{We adopt the HQ version by default and specify the version otherwise}.


\input{table/tb_datasets}




\vspace{-8pt}
\paragraph{Implementation.}
We modify Xception~\cite{8099678,rossler2019faceforensics++} as the backbone network. 
We use DLIB~\cite{sagonas2016300} for face extraction and alignment and resize the aligned faces to $256\times 256$.
Model parameters are initialized by Xception pre-trained on ImageNet.
The batch size is set to 32.
Adam~\cite{kingma2014adam} is used for optimization with a learning rate of 0.0002.
Details of the two-stream network structure and training settings are presented in the supplementary material.



\input{table/1-ablation_study}

\subsection{Ablation Study}
To demonstrate the benefit of each module, we evaluate the proposed model and its variants on the FF++ database.
All models are trained on F2F and examined on all four datasets.
The results are presented in Tab.~\ref{tab:ablation}. 
RGB represents the Xception baseline with RGB images as input.
SRM denotes the modified model with the input image replaced by SRM noise residuals. 
Two-stream Fusion represents the basic two-stream model, which consists of an RGB stream and a high-frequency noise stream and adopts the attention-based fusion mechanism. 
RSA, DCMA, and Multi-scale represent the residual guided spatial attention module, the dual cross-modality attention module, and the multi-scale high-frequency feature extraction module.

From this experiment we get the following observations.
Firstly, compared with RGB, SRM achieves better performance on FS and NT.
Furthermore, the basic two-stream model outperforms both RGB and SRM on all datasets, indicating the two modalities' complementary nature.
As presented in the last three rows, the model's performance gradually improves with each module added step-by-step, demonstrating each module's effectiveness.


\input{table/2-cmp_on_ff}
\input{table/3-cross-db-test}

\subsection{Generalization Ability Evaluation}
To fully evaluate the proposed model's generalization ability, we conduct extensive cross-database evaluations in two different settings.
The model is compared against two competing methods, Xception~\cite{rossler2019faceforensics++} and Face X-ray~\cite{li2019face}. 
Face X-ray attempts to detect the fusion boundary brought by the blending operation.
We implement it rigorously following the companion paper's instructions and train these models under the same setting.

\vspace{-10pt}
\paragraph{Generalize from one method to another.}
We conduct this experiment on the FF++ (HQ) database~\cite{rossler2019faceforensics++} that contains forged images from four different manipulation techniques, 
\textit{i.e.,} DeepFakes (DF)~\cite{deepfake}, Face2Face (F2F)~\cite{thies2016face2face}, FaceSwap (FS)~\cite{faceswap}, and NeuralTextures (NT)~\cite{thies2019deferred}.
We use forged images of one method for training and those of all four methods for testing. 
As shown in Tab.~\ref{tab:cmp-ff}, our model exceeds the competitors in most cases.
Since Xception overly relies on the texture patterns, its performance drops drastically in unseen forgeries.
Face X-ray achieves a relatively better generalization ability as it detects the blending evidence.
Our model leverages both textures and noises and captures the blending effects in the noise space, therefore generalizing better from one method to another. 

\vspace{-8pt}
\paragraph{Generalize from FF++ to other databases.}
In this experiment, we train models on FF++ (HQ)~\cite{rossler2019faceforensics++} and evaluate them in DFD~\cite{dfd}, DFDC~\cite{dfdc}, CelebDF~\cite{li2019celeb}, and DF1.0~\cite{deeperforensics10}, respectively.
This setting is more challenging than the first one since the testing sets 
share much less similarity with the training set.
We can see from Tab.~\ref{tab:cmp-unseen} that our method achieves apparent improvements over Xception and Face X-ray on all the databases, especially on DFD, DFDC, and CelebDF.
This is because Face X-ray learns to identify the boundary patterns that are sensitive to the post-process operations varying in different databases.
On the contrary, our model learns more robust representations.

We present more statistics on the cross-database evaluation and the comparison result with Face X-ray on the blend face~(BI) dataset in the supplementary material.



\subsection{Comparison with recent works}
\paragraph{Comparison with methods utilizing high-frequency features.} 
We first compare with methods exploiting high-frequency features, \textit{i.e.}, SRMNet~\cite{zhou2018learning}, Bayar Conv~\cite{10.1145/2909827.2930786}, SSTNet~\cite{wu2020sstnet}, and F3Net~\cite{f3net}. 
The former three methods target video forges and share the same backbone but different high-pass filters, \textit{i.e.}, SRM filters, Bayar Conv filters, and loosely-constrained residual filters.
F3Net adopts Discrete Cosine Transform to estimate frequency statistics.
We follow their setting and perform video-level detection on the highly compressed (LQ) FF++ database.
Since our model is trained at the image level, we sample 1 frame every five frames to collect a total of 25 images for each video.
Then we average the predictions over sampled images to classify each video.
Results are presented in  Tab.~\ref{tab:cmp-residual}. 
Note that we compare against F3Net with the Xception backbone for a fair comparison.
Our method achieves comparable robustness with F3Net and better performance than the others.

\vspace{-8pt}
\paragraph{Comparison with methods using multi-task learning.} 

We then compare our model against several methods that adopt multi-task learning for a better generalization ability, including LAE~\cite{du2019towards}, ClassNSeg~\cite{nguyen2019multi}, and ForensicTrans~\cite{cozzolino2018forensictransfer}. 
These three methods perform forgery localization and classification simultaneously.
Note that ForensicTrans adopts image residuals as input and needs to be fine-tuned on a few samples from the target domain.
Following these methods, we train our model on F2F and test it on both F2F and FS.
The results of the competing methods are the reported statistics in the corresponding papers. 
As illustrated in Tab.~\ref{tab:cmp-generalization}, 
our method surpasses the competitors in both within-database and cross-database evaluations.

\vspace{-8pt}
\paragraph{Comparison with other state-of-the-art methods.} 
We further compare our model with FWA~\cite{li2018exposing} and FFD~\cite{stehouwer2019detection}. 
FWA focuses on post-processing artifacts such as blurring and warping effects, and it is trained on synthetic data which mimics those artifacts explicitly. 
FFD exploits annotated forgery masks to provide supervision on attention maps. 
Note that FFD is trained on the DFFD~\cite{stehouwer2019detection} database, which contains FF++ and forges from other manipulation algorithms.
We adopt CelebDF as the testing set. 
As shown in Tab.~\ref{tab:cmp-att},
our method outperforms the competing methods by more than $15\%$ in AUC. 

\input{table/5-cmp-with-residual-related}
\input{table/4-cmp-on-generalization}
\input{table/6-cmp-with-attention-related}

%% file: figure/att_example.tex
\begin{figure*}[t]
  \centering
 \includegraphics[width=0.75\linewidth]{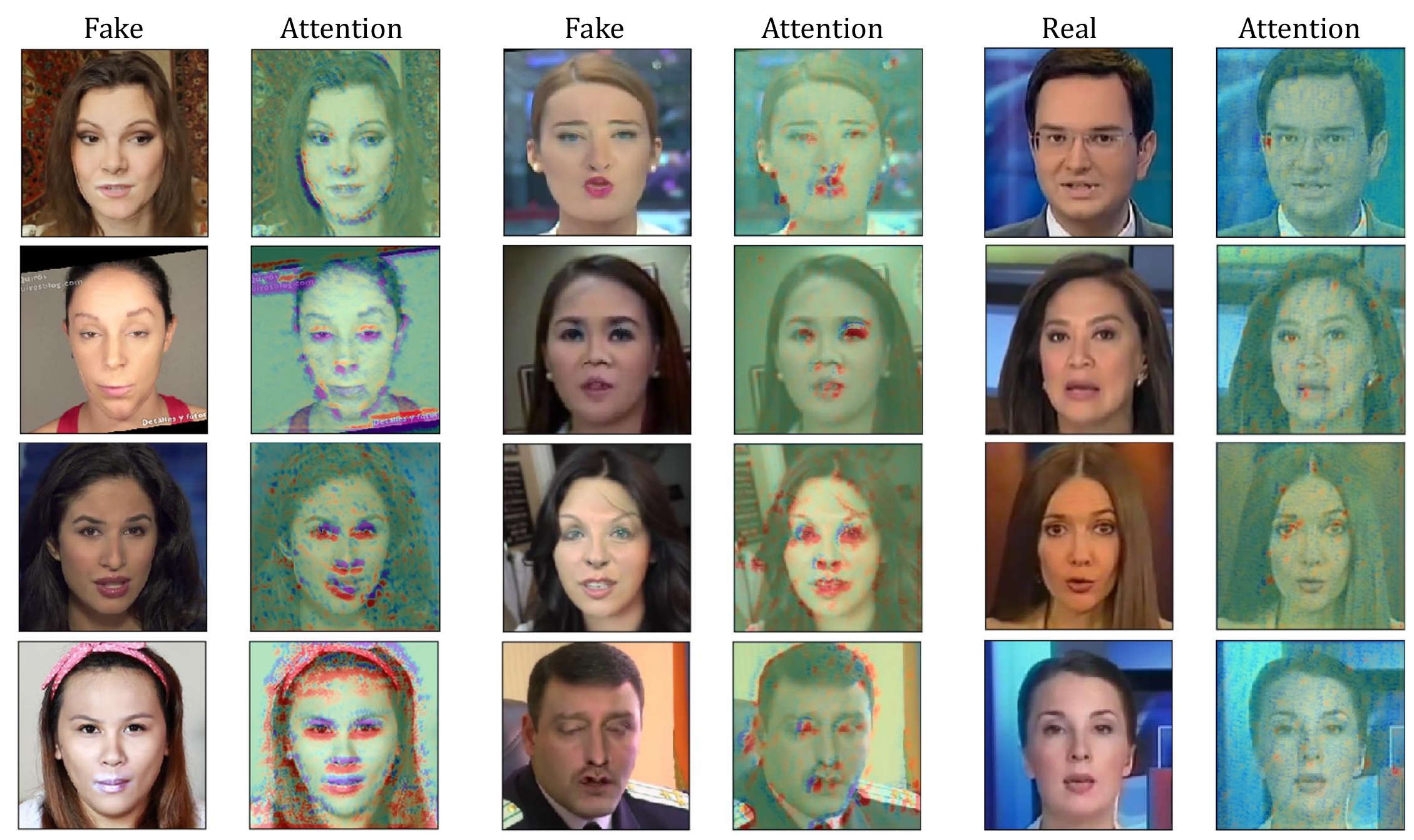}
  \caption{
  Examples of the residual guided attention.
  High responses occur around abnormal facial boundaries in manipulated faces while distribute uniformly in real ones.}
 \label{fig:att-example}
  \vspace{-0.4cm} 
\end{figure*}

%% file: table/tb_datasets.tex
\begin{table}[tb!]
  \centering
  \caption{Specifications of benchmark databases.}
  \vspace{-0.1cm}
  \scalebox{0.75}{
  \begin{tabular}{c|c|c} 
    \toprule
    Database & Video Scale &Manipulation Algorithm \\
    \midrule
    FF++~\cite{rossler2019faceforensics++}  & 1000 real, 4000 fake
  & DF~\cite{deepfake}, FS~\cite{faceswap}, NT~\cite{thies2019deferred}, F2F~\cite{thies2016face2face} \\
    DFD~\cite{dfd}      & {~~}363 real, 3068 fake
      & Improved DF \\
    CelebDF~\cite{li2019celeb}  & {~~}408 real, 795{~~} fake
   & Improved DF \\
    DFDC~\cite{dfdc}   & 1133 real, 4080 fake
   & Unpublished \\ 
   DF1.0~\cite{deeperforensics10}& 11,000 fake &DF-VAE~\cite{deeperforensics10}\\
   \bottomrule
  \end{tabular}
  }
  \label{tab:datasets}
\end{table}



%% file: table/1-ablation_study.tex
\begin{table}[tb!]
  \centering
  \caption{Ablation study on FF++. The metric is AUC. Results in gray indicate the within-dataset performance.}
  \vspace{-5pt}
  \scalebox{0.75}{
  \begin{tabular}{c|cccc}
    \toprule 
    Method
    ~& DF& F2F& FS& NT \\ 
    \midrule
    
    RGB& {0.803}& \cellcolor{Gray}0.994& 0.762& 0.696 \\
    SRM& 0.758& \cellcolor{Gray}0.994& {0.913}& {0.858} \\
    \midrule
    
    Two-stream Fusion (Fusion) & 0.810& \cellcolor{Gray}0.994& 0.922& 0.894 \\
    Fusion + RSA & 0.819& \cellcolor{Gray}0.995& 0.966& 0.927 \\
    Fusion + RSA + DCMA& 0.801& \cellcolor{Gray}0.995& 0.975& 0.957 \\
    Fusion + RSA + DCMA + Multi-scale& \textbf{0.837}& \cellcolor{Gray}0.994& \textbf{0.987}& \textbf{0.984} \\ 
    \bottomrule
  \end{tabular}}
  
  \label{tab:ablation}
  \vspace{-0.2cm} 
\end{table}

%% file: table/2-cmp_on_ff.tex
    \begin{table}[tb!]
      \centering
      \caption{
      Cross-database evaluation on FF++ database (HQ). 
      }
      \scalebox{0.80}{
      \begin{tabular}{c|c|cccc} 
        \toprule
        \multirow{2}*{Training Set} & \multirow{2}*{Model} & \multicolumn{4}{c}{Testing Set (AUC)} \\
        \cmidrule(lr){3-6}
        ~& ~& DF& F2F& FS& NT \\ 
        \midrule

        \multirow{3}*{DF}
        &Xception~\cite{rossler2019faceforensics++}& \cellcolor{Gray}\textbf{0.993}& 0.736& 0.490& 0.736 \\
        ~& Face X-ray~\cite{li2019face}& \cellcolor{Gray}0.987& 0.633& \textbf{0.600}& 0.698 \\
        ~& Ours& \cellcolor{Gray} 0.992& \textbf{0.764}& 0.497& \textbf{0.814} \\
        \midrule

        \multirow{3}*{F2F}
        &Xception~\cite{rossler2019faceforensics++}& 0.803& \cellcolor{Gray}\textbf{0.994}& 0.762& 0.696 \\
        ~& Face X-ray~\cite{li2019face}& 0.630& \cellcolor{Gray}0.984& 0.938& 0.945 \\
        ~& Ours& \textbf{0.837}& \cellcolor{Gray}\textbf{0.994}& \textbf{0.987}& \textbf{0.984} \\
        \midrule
	
        \multirow{3}*{FS} 
        &Xception~\cite{rossler2019faceforensics++}& 0.664& 0.888& \cellcolor{Gray}0.994& 0.713 \\
        ~& Face X-ray~\cite{li2019face}& 0.458& 0.961& \cellcolor{Gray}0.981& 0.957 \\
        ~& Ours& \textbf{0.685}& \textbf{0.993}& \cellcolor{Gray}\textbf{0.995}& \textbf{0.980} \\
        \midrule

        \multirow{3}*{NT}
        &Xception~\cite{rossler2019faceforensics++}& 0.799& 0.813& 0.731& \cellcolor{Gray}0.991 \\
        ~& Face X-ray~\cite{li2019face}& 0.705& 0.917& 0.910& \cellcolor{Gray}0.925 \\
        ~& Ours& \textbf{0.894}& \textbf{0.995}& \textbf{0.993}& \cellcolor{Gray}\textbf{0.994} \\
        \bottomrule
      \end{tabular}}
      \label{tab:cmp-ff}
      \vspace{-5pt} 
    \end{table}

%% file: table/3-cross-db-test.tex
        
        

    
\begin{table}[tb!]
      \centering
      \caption{
      Cross-database evaluation from FF++ to others. 
    }
      \scalebox{0.8}{
      \begin{tabular}{c|c|c|c|c|c}
        \toprule
        \multirow{2}*{Training} & \multirow{2}*{Model}
        &\multicolumn{4}{c}{Testing AUC}\\
        \cmidrule(lr){3-6}
        ~&~&
        DFD &DFDC &CelebDF& DF1.0 \\
        \midrule
        
        \multirow{3}*{FF++}
        & Xception~\cite{rossler2019faceforensics++}
        & 0.831&  0.679&  0.594& ~ 0.698\\
        
        ~& Face X-ray~\cite{li2019face}& 0.856& 0.700&  0.742& ~ 0.723\\

        ~& Ours& \textbf{0.919}&  \textbf{0.797}&  \textbf{0.794}& ~ \textbf{0.738}\\
         \bottomrule
      \end{tabular}}
      \label{tab:cmp-unseen}
      \vspace{-10pt} 
    \end{table}

%% file: table/5-cmp-with-residual-related.tex
    \begin{table}[tb!]
      \centering
      \caption{Comparison on FF++ with methods using high-frequency features. The metric is accuracy.}
      \vspace{-5pt}
      \scalebox{0.80}{
      \begin{tabular}{c|cccc} \toprule
        \multirow{2}*{Model} &\multicolumn{4}{c}{Training/Testing Set (LQ)} \\
        \cmidrule(lr){2-5}
        ~& DF& F2F& FS& Real\\ 
        \midrule
        
        SRMNet~\cite{zhou2018learning} & 0.919& 0.927& 0.891& 0.693 \\
        Bayar Conv~\cite{10.1145/2909827.2930786} & 0.929& 0.946& 0.897& 0.755 \\
        SSTNet~\cite{wu2020sstnet} & 0.934& 0.919& 0.919& 0.793 \\
        F3Net~\cite{f3net} & 0.980& 0.953& \textbf{0.965}& - \\
        Ours & \textbf{0.986} &\textbf{0.957} &0.929 & \textbf{0.971}\\
         \bottomrule
      \end{tabular}}
      \label{tab:cmp-residual}
    \end{table}

%% file: table/4-cmp-on-generalization.tex
    \begin{table}[tb!]
      \centering
      \caption{Comparison with recent works using multi-task learning. The metric is accuracy.}
      \vspace{-5pt}
      \scalebox{0.80}{
      \begin{tabular}{c|c|cc} \toprule
        \multirow{2}*{Training Set} & \multirow{2}*{Model} &\multicolumn{2}{c}{Testing Set (Acc)} \\
        \cmidrule(lr){3-4}
        ~& ~& FS(HQ)& F2F(HQ)\\ 
        \midrule
        
        \multirow{5}*{F2F(HQ)}& 
        LAE~\cite{du2019towards}& 0.632& \cellcolor{Gray}0.909 \\
        ~& ClassNSeg~\cite{nguyen2019multi} & 0.541& \cellcolor{Gray}0.928 \\
        ~& ForensicTrans~\cite{cozzolino2018forensictransfer} & 0.726& \cellcolor{Gray}0.945 \\
        ~& Ours& \textbf{0.867} & \cellcolor{Gray}\textbf{0.992} \\
         \bottomrule
      \end{tabular}}
      \label{tab:cmp-generalization}
    \end{table}

%% file: table/6-cmp-with-attention-related.tex
    \begin{table}[tb!]
      \centering
      \caption{Comparison on CelebDF. The metric is AUC.}
      \vspace{-5pt}
      \scalebox{0.80}{
      \begin{tabular}{c|c|c} \toprule
        Model& Training Set& Testing AUC on CelebDF\\ 
        \midrule
        FWA~\cite{li2018exposing} & self-collected & 0.538 \\
        FFD~\cite{stehouwer2019detection} & DFFD~\cite{stehouwer2019detection} & 0.644 \\
        Ours & FF++(HQ)& \textbf{0.794} \\
        \bottomrule
      \end{tabular}
      }
      \label{tab:cmp-att}
    \end{table}

%% file: tex/6.conclusion.tex
\section{Conclusion}

In this paper, we conducted studies on the CNN-based forgery detector, finding that the CNN detector is easily overfitted to method-specific texture patterns.
To learn more robust representations, we proposed to utilize the image's high-frequency noise features, which remove the color textures and reveal forgery traces.
We leveraged both the color textures and the high-frequency noises and proposed a new face forgery detector.
Three functional modules were carefully devised, \textit{i.e.,} a multi-scale high-frequency feature extraction module, a residual guided spatial attention module, and a dual cross-modality attention module.
We employed the proposed modules to extract more informative features and capture the correlation and interaction between complementary modalities.
The comprehensive experiments demonstrate the effectiveness of each module, and the comparisons with the competing methods further corroborate our model's superior generalization ability.

%% file: tex/acknowledge.tex
\section*{Acknowledgement}

This work was partly supported by the National Key Research and Development Program of China (2018AAA0100704), NSFC (61972250, U19B2035), CCF-Tencent Open Fund (RAGR20200113), and Shanghai Municipal Science and Technology Major Project (2021SHZDZX0102).